\documentstyle[10pt,pacling97,examples]{article}
\makeatother		
\newcommand{\scare}[1]{`#1'}

\newcommand{\term}[1]{{\sc #1}}
\newcommand{\lingform}[1]{{\it #1}}

\exampleindent\leftmargini

\title{Reluctant Paraphrase: \\
Textual Restructuring under an Optimisation Model}

\author{Mark Dras \\
Language Technology Group, Microsoft Research Institute \\
School of MPCE, Macquarie University \\
Sydney NSW Australia 2109 \\
markd@mpce.mq.edu.au}

\begin{document}

\vspace{0.1in}

\begin{abstract}
This paper develops a computational model of paraphrase under which text
modification is carried out reluctantly; that is, there are external
constraints, such as length or readability, on an otherwise ideal text, and modifications
to the text are necessary to ensure conformance to these constraints.
This problem is analogous to a mathematical optimisation problem: 
the textual constraints can be described as a set of constraint equations,
and the requirement for minimal change to the text can be expressed as
a function to be minimised; so techniques from this domain can be used to
solve the problem.

The work is done as part of a computational paraphrase
system using the XTAG system \cite{xtag} as a base.  The paper will 
present a theoretical computational framework for working within the
Reluctant Paraphrase paradigm: three types of textual constraints are specified,
effects of paraphrase on text are described, and a model incorporating
mathematical optimisation techniques is outlined.
\end{abstract}

\maketitle



\section{Framework} \label{framework}

The work this paper describes is done as part of a computational paraphrase
system using the XTAG system \cite{xtag} as a base.  Although the goal of 
the system is to modify text to achieve some objective, it is fundamentally
unlike existing systems which paraphrase text, such as style checkers \cite{kieras},
in that the context of paraphrasing is different; this context, Reluctant
Paraphrasing, is described below, with a theoretical framework for the
paraphrasing presented in the rest of the paper.

Reluctant Paraphrase (RP) can best be defined by contrasting it with the
remedial sort of paraphrases suggested by style checkers,
or in style guides such as Strunk and White \cite{strunk}, and so
on.  The starting point under this remedial style of paraphrase is an imperfect text which has
to be corrected, the corrections being determined by some prescriptive
advice such as ``make the text more
active''.  The text is run through a style checker, or past an editor, and flaws of
vocabulary or grammar or style are corrected.  In contrast, imagine the completion
of an ideal document: it says exactly what the author intends, and no more; every
word captures all the nuances the author wants to convey.  However, it has to be 
changed because of external constraints.
These constraints might be the need to cut down an academic paper by one
page for conference publication; or the need to make a technical document
conform to house style readability requirements;
or some combination of these or other sorts of external
constraints.  Thus, the text has to be paraphrased, albeit reluctantly, in order 
to meet these externally imposed constraints.

Dealing with this reluctant sort of paraphrase, rather than the remedial sort, has a
number of advantages.  Firstly, it avoids representational
problems that are otherwise inherent in paraphrasing.  In remedial paraphrasing,
paraphrase requirements can be of arbitrary complexity, ranging from ``change
sentence voice'' to ``fix incoherent theme''.
This arbitrariness of complexity makes developing a consistent representation
near impossible.  However, under RP the paraphrases don't embody the 
correction in the same way that remedial paraphrases do; instead, they are just tools
which are used to alter the text so that it conforms to the imposed constraints. 
Given that 
the paraphrases are just tools, it is possible to pick a limited set of them and still
attempt to cover all of them with a consistent representation.

Secondly, it avoids the debate about making text \scare{better}.  There are
longstanding arguments 
in the literature about particular techniques and their efficacy in improving
text: examples are the passive to active voice paraphrase, relative pronoun deletion
and the avoidance of nominalisation.
In RP, by contrast, taking the standpoint that the original
text is ideal means that any change will be undesirable, so only the minimal level of
change to the text in keeping with the constraint satisfaction should be made.

The computational paraphrase
system within RP that this paper discusses
thus has three components: a set of paraphrase
techniques which is used to achieve the text modification; a set of constraints
to which the text must adhere after the modification; and an 
effect---that of the change to the text caused by the paraphrases applied---which
is to be minimised. 
This parallels closely a mathematical optimisation model, with, respectively,
a set of decision variables, a set of constraint
equations and an optimisation function.  The rest of this paper 
presents a formulation of RP which draws on ideas from the field of
mathematical optimisation:
Section~\ref{constraints} discusses numeric constraints on text; 
Section~\ref{par} looks at quantifying text effects of paraphrases; and
Section~\ref{opt} describes the actual model.

\section{Textual Constraints} \label{constraints}

This section describes three measures of text, those of length, readability
and lexical density.  These measures are often used in the production of
text; their numeric quality is what makes them particularly amenable to
the optimisation model of this paper.

Length is the simplest measure, and is frequently used in practice as a constraint.
For example, restricting the length of a text is standard for academic 
conferences---like this conference with its 3000 word limit on abstracts---and
meeting this constraint often involves cutting down a longer draft version.
It is also typical in other areas such as the editing of newspaper text \cite{bell}.
Constraining text length is also a feature of computational language
generation systems, either as a general directive implementing the Gricean
maxim of conciseness, as in the Epicure system \cite{dale}, or as an explicit
limit on the length of an individual text unit, as in the {\sc Streak} system 
\cite{robin}.

Another common measure comes from readability formulae,
such as the Flesch Reading Ease Score or the Dale-Chall formula \cite{klare}.
Standard readability formulae are basically equations which attempt to {\em predict},
rather than evaluate, the readability of text; in form they are generally linear
combinations of factors which correlate with text complexity.  These
factors are of fairly simple types: a measure of sentence complexity, usually
average sentence length; and a measure of word complexity, such as average
word length in syllables, or proportion of infrequent words.  The weightings
for these terms are assigned by calculating a correlation with tests of
readers' comprehension.

The most accurate way of determining readability would be by testing
readers' comprehension directly.  However, this would be expensive in terms
of time and other resources; readability formulae were constructed as an
attempt to predict the readability that would be measured by these tests.  This,
together with the numerical phrasing of the readability, is the reason for using
readability formulae here.  
Moreover, the faults of readability formulae---documented in, for example,
\cite{duffy}---are not significant in the context of RP, for a number
of reasons.

Firstly, use of readability formulae can be defended on practical 
grounds: readability
formulae are used as criteria for writing public documents in the US, such as
insurance policies, tax forms, contracts and jury instructions \cite{bruce},
for producing military documents 
\cite{kincaid}, and so on.  In these situations the use
of readability formulae is mandatory; so for a system which models realistic
constraints on text, using the formulae as a constraint is reasonable.

Secondly, most objections are based on the use of readability formulae in the
strong sense---when
actual readability levels are predicted---rather than when used in their
weak sense---when readability formulae are used to rank texts relative
to each other in order of reading complexity \cite{kern}; 
and under Reluctant Paraphrase,
this is not a problem, as the texts, one of which is a paraphrase
of the other, are just ranked relative to each other.

Lexical density is a textual measure discussed by Halliday \cite{halliday:b}; 
it attempts to
capture the \scare{condensedness} of text by measuring the proportion of
non-content (or function) words to total text.  Halliday uses this idea of
condensedness to distinguish between written and spoken forms of language:
written language tends to be more condensed than spoken, with constructions
of type (\ref{ld}a) more prevalent in writing and those of type (\ref{ld}b)
more prevalent in speech.

\begin{examples}
\item \label{ld}
\begin{subexamples}
\item Sex determination varies in different organisms.
\item The way sex is determined varies in different organisms.
\end{subexamples}
\end{examples}

The concept is also useful in the context of this paper's optimisation model, as a
constraint counterbalancing the readability one.  
Under a typical readability formula,
the readability value is generally correlated with average sentence length, so the formula value
can be improved by the sort of paraphrases which compress text, such as the
mapping of (\ref{ld}b) to (\ref{ld}a).  Compression to too great an extent can
lead to text that is difficult to understand; the use of lexical density as a constraint
can act as a counterweight to the readability constraint, to prevent excessive
text compression.

\section{Paraphrases} \label{par}

As noted in Section \ref{framework}, paraphrases can be of arbitrary complexity.  In
keeping with their use in RP as broad-coverage tools, the
most appropriate paraphrases, and hence the ones that are used in this work,
are ones that are syntactic in nature.
An example of this type, modelled on work by Jordan \cite{jordan}, is the
splitting off of a noun
post-modifier to form a separate sentence:

\begin{examples}
\item
\begin{subexamples}
\item Sarah warily eyed the page filled with topicalisations and other
linguistic phenomena.
\item Sarah warily eyed the page.  It was filled with topicalisations and other
linguistic phenomena.
\end{subexamples}
\end{examples}

The paraphrases used here are taken from three different types of sources:
popular (style guides such as \cite{strunk});
academic (work on textual analysis involving paraphrasing, 
such as \cite{jordan} and \cite{robin});
and practical (the actual practices of people involved in paraphrasing text,
such as editors and journalists \cite{bell}).

These paraphrases will cause some change to the text, and, under RP, any
change effected by a paraphrase is taken to be a negative one.
Developing an optimisation model thus requires a quantification of the effects that
imposing a paraphrase on a text will have on that text.  The rest of this
section sketches
methods for assigning a quantification to a paraphrase, which will lead to a
minimisation function for the model.  There are two types of effects analysed in
this work, effects on meaning and effects on discourse structure.  These two
types are then combined to give the minimisation function.

\subsection{Meaning Effects}

One way in which a paraphrase can affect a text is in terms of its truth-conditional
meaning; or, in Hallidayan terms, its ideational metafunction.  A unit of text, such
as a sentence, can be viewed as a statement about the world, which is either true
or false\footnote{Only declarative sentences are dealt with in this paper.}; 
an alternative, but related, view is that the truth of the statement is
represented by a set of possible worlds in which the statement is true\footnote{This
is a much simplified summary of work on truth-conditional meaning presented
in, for example, \cite{allwood}.}.
A paraphrase is consequently defined more precisely as consisting
of two sentences where the set of possible worlds in which one sentence is true is a
(not necessarily proper) subset of the possible worlds in which the other is true.
Take the following examples:

\begin{examples}
\item \label{meaning_ex1}
\begin{subexamples}
\item Onlookers scrambled to avoid the car which was flashing its headlights.
\item Onlookers scrambled to avoid the car flashing its headlights.
\end{subexamples}
\item \label{meaning_ex2}
\begin{subexamples}
\item The salesman made an attempt to wear Steven down.
\item The salesman attempted to wear Steven down.
\end{subexamples}
\item \label{meaning_ex3}
\begin{subexamples}
\item There was a girl standing in the corner.
\item There was a girl in the corner.
\end{subexamples}
\item \label{meaning_ex4}
\begin{subexamples}
\item Tempeste approached Blade, a midnight dark and powerful figure,
and gave him a resounding slap.
\item Tempeste approached Blade and gave him a resounding slap.
\end{subexamples}
\end{examples}

These examples give a range of different magnitudes in the size of the sets
representing the possible worlds in which each of the paraphrase alternatives
is true.  Example (\ref{meaning_ex1}) represents a fairly minimal difference:
(\ref{meaning_ex1}b) can be a paraphrase either of (\ref{meaning_ex1}a) or of
\lingform{Onlookers scrambled to avoid the car which {\em is} flashing its headlights}.
The possible worlds in which (\ref{meaning_ex1}b) is true
is a proper superset of the possible worlds in which (\ref{meaning_ex1}a) is
true; but intuition suggests the sets are relatively close in size,
(\ref{meaning_ex1}b) only covering two different cases with respect to the
altered constituents.  Example (\ref{meaning_ex2}) represents a slightly bigger
paraphrase: (\ref{meaning_ex2}b) can paraphrase statements asserting one
attempt---equivalent to (\ref{meaning_ex2}a)---two attempts, seven attempts,
or many attempts.  The size of the set difference here is consequently 
relatively larger
than in (\ref{meaning_ex1}).  In (\ref{meaning_ex3}), the difference is
larger still, in that (\ref{meaning_ex3}b) can describe situations where the girl
is sitting, lying, dancing, and so on.  The largest difference is in
(\ref{meaning_ex4}), where (\ref{meaning_ex4}b) includes in its set of
possible worlds, over and above the possible worlds in which (\ref{meaning_ex4}a)
is true, worlds in which Blade is described by any other appositive.

A way of approximating the intuition about the difference in the relative sizes
of possible world sets is by using parts of speech.  An alteration in less
significant parts of speech corresponds to a small relative difference in set size, 
and so on.  So in (\ref{meaning_ex1}), the changed parts of speech are a relative
pronoun, which causes no difference in truth-conditional meaning, and the
auxiliary verb \lingform{be}, which leads to the relatively small difference.  In
comparison, the deletion of the open-class constituent in ({\ref{meaning_ex3}),
the present participle \lingform{standing}, leads to a much greater set difference;
and deleting multiple open-class words in (\ref{meaning_ex4}) has a still larger
effect.

A possible refinement of this approximation involves considering lexical 
factors.  For example,
the paraphrase in (\ref{meaning_ex3}) is less significant than if (\ref{meaning_ex3}a)
had been \lingform{the girl coruscating in the corner}; the latter option is much more
unexpected, and so it can be argued that its removal alters the text to 
a much greater extent.  As they are related to frequency, these
lexical factors could be estimated through collocational analysis within a
corpus, although this has not been done as yet.

\subsection{Discourse Effects}

As well as affecting the truth-conditional meaning of the text, a paraphrase can
alter the discourse features of the text; or, in Hallidayan terms again, the textual
metafunction.
Because of the assumption behind RP that the author has
deliberately chosen a particular way of packaging the information in a sentence,
any paraphrase which alters the packaging structure is altering the author's
intention and hence should be included in the measurement of change and the
consequent minimisation function.  
Work in the area of information packaging includes \cite{givon}, 
\cite{halliday:a} and
\cite{vallduvi}; 
although approaches differ, all have some concept of
syntactic structures reflecting packaging of information---which part is known to
the reader, and which is new.
An example is an \lingform{it}-cleft
sentence and its standard declarative paraphrase:

\begin{examples}
\item \label{discourse_ex1}
\begin{subexamples}
\item It was the balcony and its scholarly discourse which irresistibly drew Ryan.
\item The balcony and its scholarly discourse irresistibly drew Ryan.
\end{subexamples}
\end{examples}

In (\ref{discourse_ex1}a), the fact that Ryan has been irresistibly drawn is
indicated as a given or topic, and the balcony-as-drawer
as the new piece of information.
In (\ref{discourse_ex1}b) there is no such marking.

A rough numerical measure of this can be gained by counting the difference in
the questions to which the sentence can be an answer.  So (\ref{discourse_ex1}a)
can only be an answer to the narrow-focus \lingform{What irresistibly drew Ryan?},
while (\ref{discourse_ex1}b) can answer not only this question but also
\lingform{What did the balcony and its scholarly discourse do to Ryan?},
\lingform{What did the balcony and its scholarly discourse do?}, or the wide-focus
\lingform{What happened?}.

\section{An Optimisation Approach} \label{opt}

The optimisation model for the computational paraphrase system
requires a formal specification of the paraphrases and
their attributes---their effect on the text in terms of the parameters, such as
number of words or sentences, affected by each
constraint; and their effect on the text's meaning and information structure.  
The paraphrases are formally
specified using the representation formalism 
as proposed in \cite{dras}; however,
an informal description of the paraphrase is adequate for discussion of the
paraphrase effects and their inclusion into the optimisation model.

This section presents a mathematical optimisation model of paraphrasing.  The
basic techniques are those of integer programming (see, for example, \cite{winston}),
which describes the constraints and function to be minimised in terms of linear
combinations of integer variables.  
The integer programming approach is useful because it provides a set of
techniques for guaranteeing an optimal solution, heuristics for cutting the
search space, and methods for model analysis\footnote{This last feature is not
discussed in this paper.}.  After a formal
presentation of the model, an example is given for clarification.

\subsection{The Model}

In developing an optimisation model, it is first necessary to identify the
\term{decision variables}: that is, those factors about which a decision is to be
made.  In this case, it is the paraphrase mappings: for each paraphrase, the decision
is whether this paraphrase should be applied to the text to move it towards satisfying
the constraints while minimally perturbing the text.  In this situation, the choice
is binary, whether or not to apply the paraphrase.  Given this, the decision variables
are

\begin{quote}
$p_{ij}$ =  a 0/1 valued variable representing the 
$j$th potential paraphrase for sentence $i$
\end{quote}

The \term{objective function}, the function to be optimised, is, for RP, 
a measure of the change to the text, as described in Section~\ref{par}.   With
$c_{ij}$ being the effect (or cost) of each paraphrase, if applied, this function has
the form

\begin{quote}
\[ z = \sum c_{ij}.p_{ij} \]
\end{quote}

The constraints take the form ``total length must be decreased by at least some
constant value'', or ``readability value must be no greater than some constant value''.
Expressed mathematically, the length constraint is

\begin{quote}
\[ \sum w_{ij}.p_{ij} \leq k_1 \]
\end{quote}

where

\begin{quote}
$w_{ij}$ = change to length of sentence $i$ caused by paraphrase $ij$ \\
$k_1$ = required change to the length of text in words; $k_1 \leq 0$
\end{quote}

A simplified readability constraint\footnote{This
simplification means that non-linear, quadratic programming techniques do not have to be
introduced at this stage.}, using only the average sentence length component, is

\begin{quote}
\[ \frac{W + \sum w_{ij}.p_{ij}}{S + \sum s_{ij}.p_{ij}} \leq k_2 \]
\end{quote}

that is,

\begin{quote}
\[ \sum (w_{ij} - k_2.s_{ij})p_{ij} \leq k_2S - W \]
\end{quote}

where

\begin{quote}
$s_{ij}$ = change to number of sentences in the text by paraphrase $ij$ \\
$W$ = total number of words in original text \\
$S$ = total number of sentences in original text \\
$k_2$ = required average sentence length\footnote{While the choice
of a particular $k_1$ is straightforward, choosing a reasonable value for
$k_2$ requires more effort: for example, analysing average sentence length
in a corpus which satisfies typical readability targets (such as ``senior
high school level" in the Flesch Reading Ease score).  The constant $k_3$
can be ascertained similarly.}; $k_2 \geq 0$
\end{quote}

The lexical density constraint requires the proportion of function words, taken
here to be all closed class words, to total words to be greater than some constant
value.  It has the form

\begin{quote}
\[ \frac{F + \sum f_{ij}.p_{ij}}{W + \sum w_{ij}.p_{ij}} \geq k_3 \]
\end{quote}

that is,

\begin{quote}
\[ \sum (f_{ij} - k_3.w_{ij})p_{ij} \geq k_3W - F \]
\end{quote}

where

\begin{quote}
$f_{ij}$ = change to number of function words caused by paraphrase $ij$ \\
$F$ = total number of function words in original text \\
$k_3$ = required proportion of function words to total words; $0 \leq k_3 \leq 1$
\end{quote}

Given that there are $j$ paraphrases for each sentence (with $j$ varying for
each sentence), there is a potential conflict for the paraphrases.  To simplify the
application of the paraphrases, an extra constraint is added, stating that there
can be at most one paraphrase for each sentence:

\begin{quote}
\[ \sum_j p_{ij} \leq 1 \]
\end{quote}

Although it is possible in particular cases for paraphrases to overlap and
produce satisfactory text, there is no easy way in advance to decide this;
so for an automated system the above constraint is necessary, at least until a much
more detailed analysis of paraphrase interaction has been carried out.

An example is presented in the next section, to illustrate the model.  The
small size of this example does not allow a real demonstration of the
usefulness of the approach, since the problem can be solved almost by
inspection.  However, in larger problems this method of modelling 
allows the use of techniques such as branch-and-bound \cite{winston} which
make the solution of the problem feasible, where the solution would otherwise be
impractical because of the problem's exponential complexity.

\subsection{An Example} \label{ex_section}

As an example, take the short text:

\begin{examples}
\item
\begin{subexamples}
\item The cat sat on the mat which was by the door.
\item It ate the cream ladled out by its owner.
\item The owner, an eminent engineer, had a convertible used in a bank robbery.
\end{subexamples}
\end{examples}

The values of $F$, $W$ and $S$ are 17, 33 and 3 respectively.

\begin{table} \label{coeff_table}
\begin{center}
\begin{tabular} {|c|c|c|c|} 
\hline
paraphrase $ij$ & $f_{ij}$ & $w_{ij}$ & $s_{ij}$ \\
\hline
11 & -2 & -2 & 0 \\
\hline
21 & +3 & +3 & +1 \\
\hline
31 & +3 & +3 & +1 \\
\hline
32 & -1 & -3 & 0 \\
\hline
\end{tabular}
\caption{\bf Variable coefficients}
\end{center}
\end{table}

Possible paraphrases of individual sentences, using just relative pronoun deletion,
post-modifier split, and parenthetical deletion, are:

\begin{examples}
\item
\begin{subexamples}
\item[$p_{11}$] The cat sat on the mat by the door.
\item[$p_{21}$] It ate the cream.  It had been ladled out by its owner.
\item[$p_{31}$] The owner, an eminent engineer, had a convertible.
It had been used in a bank robbery.
\item[$p_{32}$] The owner had a convertible used in a bank robbery.
\end{subexamples}
\end{examples}


\begin{table*} \label{flex_table}
\begin{center}
\begin{tabular} {|l|c|c|}
\hline
& number of  words & avg sent. length \\
\hline
original text & 1791 & 24.88 \\
\hline
num. words minimised & 1531 & 23.92 \\
\hline
avg sent. minimised & 1784 & 17.66 \\
\hline
\end{tabular}
\caption{\bf Maximal text flexibility}
\end{center}
\end{table*}

This gives decision variables $p_{11}$, $p_{21}$, $p_{31}$, and $p_{32}$, with
associated coefficients in 
Table 1.

For the example, the constraint values are (arbitrarily) chosen as $k_1 = 0$ (at
worst no compression of text length), $k_2 = 10$ (average sentence length no
greater than 10), and $k_3 = 0.525$ (function words no less than 52.5\% of the
text).

Through the process of integer programming, there are two alternatives which
are feasible solutions:

\begin{quote}
\[ p_{11} = p_{21} = p_{31} = 0, p_{32} = 1 \]
\[ p_{31} = 0, p_{11} = p_{21} = p_{32} = 1 \]
\end{quote}

This gives two values for the objective function, $z = c_{32}$ and
$z = c_{11} + c_{21} + c_{32}$.  Since $\forall(ij) c_{ij} > 0$---under
the Reluctant Paraphrase assumption all changes involve a positive cost---the
best alternative is the first, with only the second paraphrase for sentence number
three being applied.  The resulting text is then:

\begin{examples}
\item
\begin{subexamples}
\item The cat sat on the mat which was by the door.
\item It ate the cream ladled out by its owner.
\item The owner had a convertible used in a bank robbery.
\end{subexamples}
\end{examples}

\subsection{Actual Text}

Current work involves applying this technique to actual text, taken from the
periodical \lingform{The Atlantic Monthly}.  This source was chosen as it has
reasonably complex text on which a large range of paraphrases can be applied.
The text consists of 72 sentences and totals 1791 words; there are 84 possible
paraphrases, over 45 of the sentences.  

In order to determine possible constraint
values for real text, it is first necessary to evaluate the flexibility of the text: to
what extent can the length be altered, say, or the readability changed?  Choosing
sets of paraphrases which maximise the relevant constraint, regardless of the
value of the cost function or the effect on other constraints, the results
given in 
Table 2 were obtained.

So at best it is possible, for this text, to reduce word length by about
15\%, and the average sentence length by about 30\%.  This information is then
used to set reasonable constraint limits.

One way in which the task of applying the model to actual text
is more complicated than the example is in the need to set
numeric values for the objective function coefficients.  In the example, because
of the small number of objective function coefficients, it is generally possible to
just compare the result of the function algebraically. 
Taking as a first attempt at a numeric objection function the assignment
of constant
differences between the classes of textual change described in 
Section~\ref{par}, the approach was applied to the first 19 sentences of the
\lingform{Atlantic Monthly} text.  Modelling the problem as a optimisation
one, combined with branch-and-bound techniques, reduced the search space
by 41.5\% from $2^{19}$ possible solutions to 306828 candidates.

\section{Conclusion}

The paper has drawn on diverse areas of linguistics and mathematics to
present a nonetheless fairly natural view of paraphrase as a mathematical
optimisation problem.  This phrasing of paraphrase as an optimisation problem
has three main components.  Firstly, three appropriate constraints have
been chosen and modelled as constraint equations.  Secondly, a method for
quantifying the effects of paraphrase on text, and their expression as an
optimisation objective function, has been discussed.  Thirdly, the model has
been described with an application to a small example text given.  Application
to actual text has shown the extent to which the technique can be applied: for
example, the length constraint is not meant to mimic summarisation, but rather
to enable the massaging of a text that is not too far from what is required.

Current work involves a deeper application of the model to actual text: a
larger number of constraints, more paraphrases, and an objective function 
which can be
numerically evaluated.  This then enables an analysis of text using the
sensitivity analysis which is a corollary of linear programming, answering
questions such as:

\begin{itemize}
\item What are the characteristics of \term{elastic} text, that is, one
which responds a lot to small changes?
\item What is the sensitivity of text to changes in model assumptions, and
would the same paraphrases be chosen given these changes?
\item What are the equivalence classes for the paraphrases used, that is,
which paraphrases are in effect interchangeable?
\end{itemize}

\end{document}